\documentclass[conference]{IEEEtran}
\IEEEoverridecommandlockouts
\usepackage{cite}
\usepackage{amsmath,amssymb,amsfonts}
\usepackage{algorithmic}
\usepackage{graphicx}
\usepackage{textcomp}
\usepackage{xcolor}
\usepackage{float}
\usepackage{caption}
\usepackage{soul}
\begin{document}

\title{FedAvg for HAR: Exploring the Tradeoff Between Personalized and Generalization Accuracy\\
\thanks{©2026 IEEE. Personal use of this material is permitted. Permission from IEEE must be obtained for all other uses. This manuscript has been accepted for publication and oral presentation at the 2026 11th IEEE International Conference on Machine Learning Technologies (ICMLT 2026). The final version will appear in IEEE Xplore.}
}

\author{\IEEEauthorblockN{Andrea De Luna}
\IEEEauthorblockA{\textit{DiSPeA} \\
\textit{University of Urbino}\\
a.deluna1@campus.uniurb.it}
\and
\IEEEauthorblockN{Susanna Peretti}
\IEEEauthorblockA{\textit{DiSPeA} \\
\textit{University of Urbino}\\
s.peretti@campus.uniurb.it}
\and 
\IEEEauthorblockN{Chiara Contoli}
\IEEEauthorblockA{\textit{DiSPeA} \\
\textit{University of Urbino}\\
chiara.contoli@uniurb.it}
\and
\IEEEauthorblockN{Alessandro Bogliolo}
\IEEEauthorblockA{\textit{DiSPeA} \\
\textit{University of Urbino}\\
alessandro.bogliolo@uniurb.it}
}

\maketitle

\begin{abstract}
The federated learning (FL) paradigm fosters distributed pervasive computing combined with artificial intelligence techniques, allowing for optimized data usage and improved mitigation of privacy concerns. Indeed, model training occurs on the client’s local devices, and model parameters are subsequently shared with a centralized server. However, there is a need to find a tradeoff between models’ personalization and generalization capabilities. In this paper, we design and implement several testing scenarios devoted to evaluating and comparing the centralized, local, and federated paradigm performances. We also design and implement a scenario that emulates a change in clients' data. We then present experimental results of the FedAvg algorithm applied to the Human Activity Recognition
(HAR) domain to understand the trade-off between personalized and generalized accuracy. Results show that, although FedAvg confirms a higher degree of personalization capabilities while keeping a high degree of generalization with respect to the traditional centralized learning, this result is not so obvious under stressful conditions, such as when varying class distribution over clients.
\end{abstract}


\section{Introduction}
Federated learning (FL) gained momentum for its ability to enhance data privacy and promote the use of devices in a distributed, pervasive computing context. Indeed, in such a scenario, clients participating in the federation collect data from their local devices, train a model locally,  and share only model parameters with a global (centralized) server. The server aggregates the model’s parameters according to a specific aggregation algorithm strategy and subsequently shares parameter updates with all the clients until a (global) convergence is met. There is a need to find a tradeoff between models’ personalization and generalization capabilities. Such a tradeoff is challenged by multiple factors, such as, but not limited to, device heterogeneity and imbalanced data distribution.
In this paper, we delve into FedAvg federated learning mechanisms applied to the Human Activity Recognition (HAR) domain to understand the trade-off between personalized and generalized accuracy. The idea is to examine the extent to which clients benefit from the FL engagement procedure compared to a non-federated scenario. Ek et al. were the pioneers in exploring the performance of federated learning model aggregations in the human activity recognition domain, where they investigated and compared FedAvg, FedPer, and FedMa \cite{ek2020evaluation}. This work was subsequently extended in \cite{sannara2021federated, ek2022evaluation}, where the authors proposed a new aggregation algorithm and compared their performance against other approaches. Many other works propose new aggregation algorithms and compare performance \cite{tu2021feddl, concone2022federated,cheng2023protohar}, some also explore how to balance accuracy and efficiency with privacy \cite{kabir2025federated}, others how to handle nonindependent and identically distributed (Non-IID) data \cite{wen2024enhancing}. 

Our work is inspired by Ek et al. \cite{ek2022evaluation} and Wen et al. \cite{wen2024enhancing}. In particular, we seek to deepen our understanding of federated algorithm aggregation dynamics. Compared to existing literature, we propose and explore two additional evaluation strategies, which are discussed in the next Section. Additionally, we explore the effect of varying, in a round robin fashion, the activity classes known to each client participating in the federation. The goal is to examine the extent to which clients benefit from the FL engagement procedure compared to a non-federated scenario. To the best of our knowledge, the activity class exclusion strategy has not been explored in the HAR federated learning literature. Results show that, although FedAvg confirms a higher degree of personalization capabilities while keeping a high degree of generalization with respect to the traditional centralized learning, this result is not so obvious under stressful conditions, such as varying class distribution over clients.


\section{Methodology}\label{sec:method}
Federated learning fosters a privacy-preserving distributed collaborative training across multiple clients that can be heterogeneous in terms of data distribution and collecting devices. The collaborative approach brings challenges with respect to global and personalized optimization. Global optimization is about training a global model from participating clients' data, whereas personalized optimization is about training a personalized model for each participating client. Our work extends the ones presented in \cite{ek2022evaluation,wen2024enhancing} by exploring additional metrics and testing scenarios to better evaluate the performance of the aggregation algorithms. In this work, for the sake of simplicity, we consider the traditional and well-known FedAvg algorithm, but the exploration can be carried out with other aggregation algorithms.
\begin{figure}[H]
    \centering
    \includegraphics[width=1\linewidth]{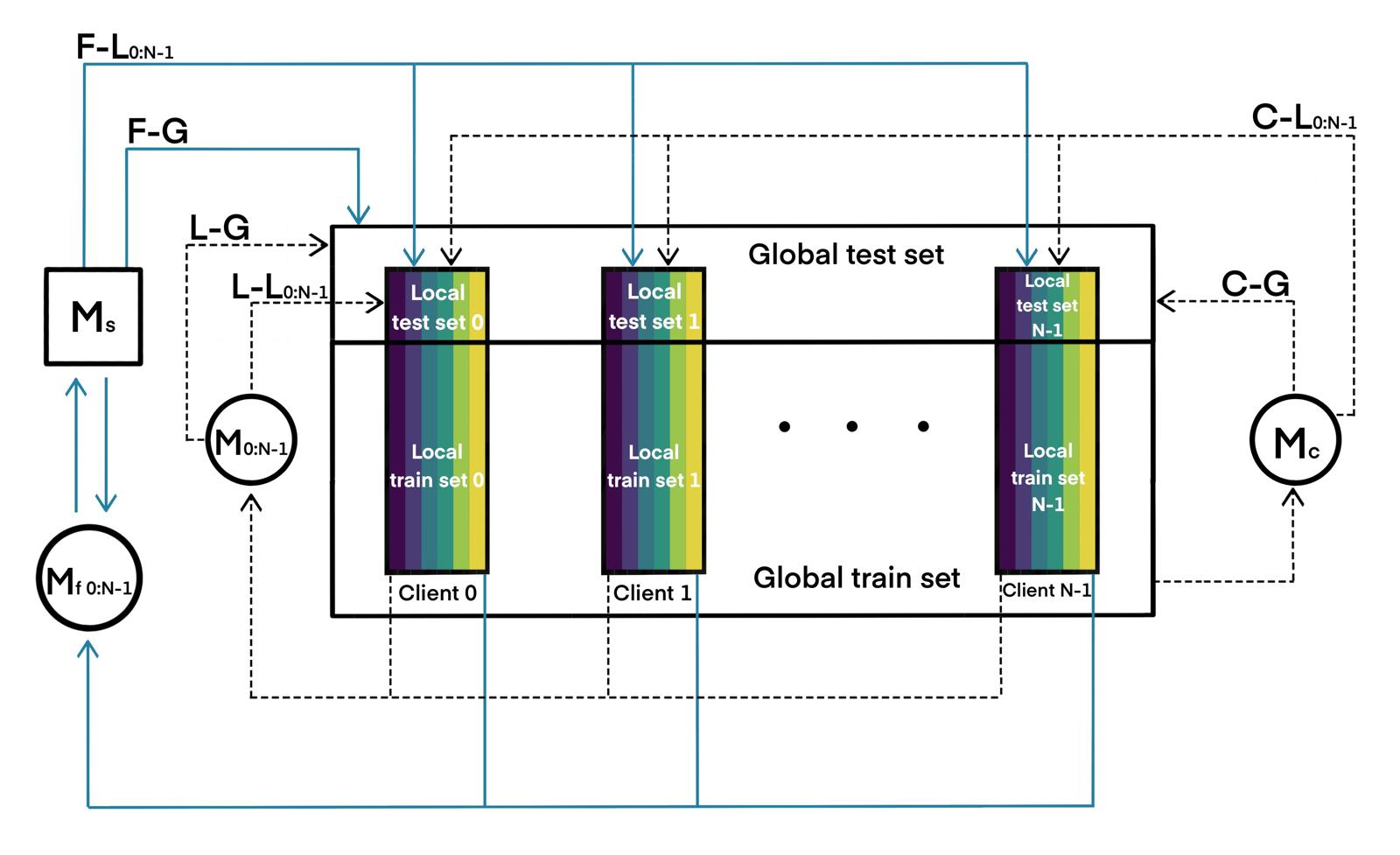}
    \caption{Data partitioning and validation scope}
    \label{fig:datapartition_metrics}
\end{figure}

\begin{table}[H]
\centering
\begin{tabular}{lll}
\hline
 & \multicolumn{2}{l}{Validation scope} \\ \cline{2-3}
Model  & Local & Global \\ \hline
Local & L-L & L-G \\
Centralized & C-L & C-G \\
PersFedAvg & PF-L & PF-G \\
FedAvg & F-L & F-G \\
\hline \\
\end{tabular}
\caption{Reference grid of models (rows) and validation scopes (columns). Each entry of the grid contains the short name used to denote the combination of models and validation options.}\label{table:model_metrics}
\end{table}

\subsection{Data Partitioning and Evaluation Metrics}\label{sec:data-partitioning}
Figure \ref{fig:datapartition_metrics}, coupled with Table \ref{table:model_metrics}, describes how we partition the dataset and which metrics we assess. A detailed description of the dataset is provided in Section \ref{sec:expsetup}. Our goal is to deepen the investigation of performance comparison between the traditional centralized approach and the federated one to foster the discussion about the tradeoff between personalization, generalization, and privacy preservation. 

First, we consider $N$ subjects, indexed from $0:N-1$,  as a client participating in the federation with its own \emph{Local train set} and its own \emph{Local test set}. Each local dataset is obtained by splitting the client's data into 80\% training and 20\% testing. The union of these training and testing sets determines \emph{Global Training Set} and the \emph{Global Test Set}, respectively. This partitioning is essential to assess the evaluation metrics. $M_{c}$ refers to the \emph{Centralized} learning model, which is trained in the traditional manner, i.e., on the global train set and tested on the Global test set. $M_{c}$ is then validated in two different scopes: in the global and local, denoted in Figure \ref{fig:datapartition_metrics} and Table \ref{table:model_metrics} as \emph{C-G} and \emph{C-L\(_{0:N-1}\)} performances, respectively. In this work, we also assess the performance of each client’s model, $M_{0:N-1}$, without considering the contribution coming from the federation, i.e., in such a scenario, the clients are not involved in the federated (algorithm) process. In this case, each client is trained on its own local train set, and it’s tested in the local scope \emph{L-L\(_{0:N-1}\)} and in the global scope \emph{L-G}. It is worth mentioning that the client’s model, in the non-federated scenario, is termed \emph{Local} in Table \ref{table:model_metrics}.  L-L is, therefore, an additional metric, which is not normally evaluated but we believe relevant to discuss the outcome of the thorough exploration.

The federated scenario is indicated by the blue lines in Figure  \ref{fig:datapartition_metrics}. According to \cite{ek2022evaluation,wen2024enhancing}, three evaluation metrics are commonly assessed in the federated context: i) \emph{global} (or \emph{aggregate}) \emph{accuracy}, ii) \emph{personalized} (or \emph{individual}) \emph{accuracy}, and iii) \emph{generalization accuracy}. The global accuracy, in Figure \ref{fig:datapartition_metrics} denoted as \emph{F-G}, is defined as the performance of the server model ($M_{s}$) trained by aggregating the contribution from all participating clients ($M_{f,0:N-1}$) to the federation. As an additional metric, we propose to validate $M_{s}$ in the local scope by evaluating \emph{F-L\(_{0:N-1}\)}. It is worth mentioning that, $M_{s}$ is termed \emph{FedAvg} in Table \ref{table:model_metrics} to indicate the specific algorithm we consider for the federated scenario. 
This evaluation is carried out after each round and is denoted in Table \ref{table:model_metrics} as \emph{F-L}. 

The personalized accuracy is defined as the performance of each client’s model ($M_{f,0:N-1}$), which is trained and tested on its local dataset, taking into account the contribution of the whole federation. For the sake of simplicity, this metric is not reported in Figure \ref{fig:datapartition_metrics}; however, in Table \ref{table:model_metrics}, we refer to this model as \emph{PersFedAvg}, and metrics as \emph{PF-L} and \emph{PF-G} for the local and global scope, respectively. The generalization accuracy is defined as the performance of each client’s model ($M_{f,0:N-1}$), which is tuned on its local dataset but is tested on a global unseen dataset. \emph{PF-L} is, therefore, in line with the definition of personalized accuracy, whereas \emph{PF-G} is in line with the definition of generalization accuracy. 

\subsection{Additional Characterization}\label{sec:additional-characterization}
In addition to the aspects discussed in Section \ref{sec:data-partitioning}, we further investigated the impact of class-level data heterogeneity within the dataset by introducing a round robin (RR) activity class removal strategy, aimed at more explicitly emulating non-independent and identically distributed (non-IID) conditions. 

This approach enforces a deterministic and uniformly distributed activity class exclusion policy across clients. Specifically, for each client, samples belonging to a single activity class are completely removed from both the local training and local testing set models. This means that each model has \emph{K} outputs, but only \emph{K-1} activities are provided as representative inputs. The activity to be excluded is selected according to a round robin rule, defined as the modulo operation between the client index and the total number of activity classes. Formally, for a client \emph{c} and a total of \emph{K} activity classes, the class to be dropped is determined as $K_{\text{drop}} = c \bmod K$. This operation is performed prior to the train–test split on each client. 
The remaining samples, referred to as seen classes, are subsequently partitioned into training and testing subsets with an 80\%–20\% split, respectively. 

During federated training, clients participate as usual using only the locally available classes, while the global model aggregates the results obtained at each communication round. In addition to the metrics defined in Section \ref{sec:data-partitioning}, this setting enables the introduction of new evaluation metrics that assess the generalization capability of the centralized model \emph{(C-)}, personalized models \emph{(FP-)}, and the final federated model \emph{(F-)} on the samples belonging to the previously excluded classes. These metrics are denoted by the suffix \emph{-LD} (Local Dropped) and are evaluated under the round robin scheme \emph{(\_RR)}.

Through this strategy, it becomes possible, first, to clearly distinguish between standard generalization performance and robustness to not-represented class scenarios, allowing for a precise evaluation of how federated and personalized learning paradigms handle partial observability. Second, it enables the assessment of the effective distributed learning capability achieved through federated learning, whereby local models are able to recognize activities that were never observed locally during training, due to the complete exclusion of the corresponding class.

\section{Experimental Setup} \label{sec:expsetup}
\subsection{Dataset}
\begin{figure}[H]
    \centering
    \includegraphics[width=0.9\linewidth]{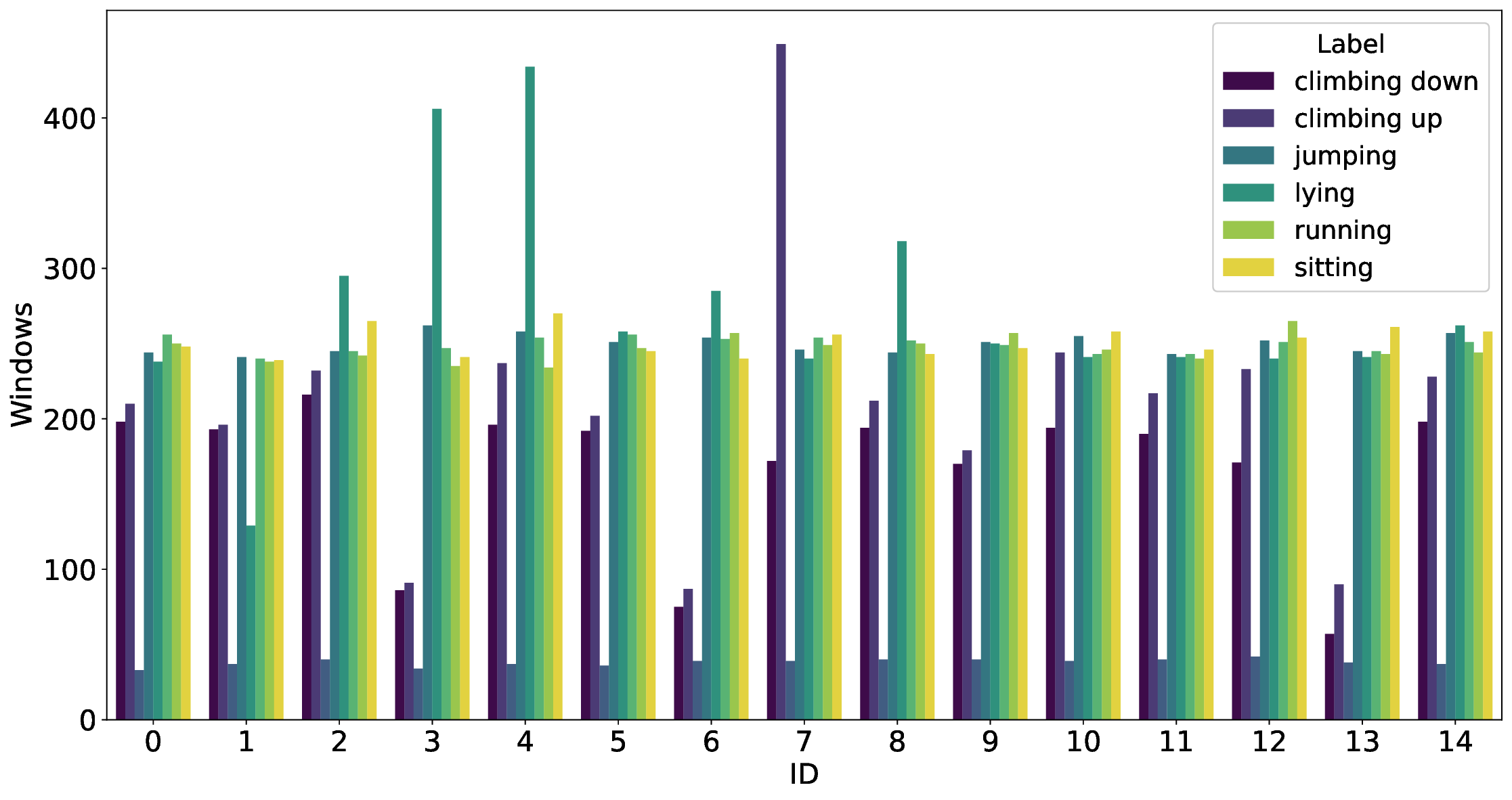}
    \caption{RealWorld2016 samples distribution for each activity across different subjects.}
    \label{fig:realw16details}
\end{figure}

The RealWorld2016 \cite{realworlddataset} dataset includes multimodal sensor data collected from fifteen subjects, consisting of eight males and seven females, with an average age of 31.9 years ($ \pm $12.4), an average height of 173.1 cm ($ \pm $6.9), and an average weight of 74.1 kg ($ \pm $13.8). It comprises measurements from acceleration, GPS, gyroscope, light, magnetic field, and sound level sensors. Each subject performed a set of activities that included climbing stairs up and down, jumping, lying, standing, sitting, running or jogging, and walking. During each activity, the acceleration was simultaneously recorded at seven body positions: the chest, forearm, head, shin, thigh, upper arm, and waist. Each activity was performed for approximately ten minutes, except for jumping, which lasted around 1.7 minutes due to its physical intensity. The data collected from male and female participants is equally balanced, and all movements were recorded on video to support analysis and annotation. Figure \ref{fig:realw16details} shows the activity distribution across subjects.

The accelerometer data were segmented using a sliding window approach with 128 samples per window. Each channel was individually z-normalized to ensure consistent scaling across signals and to improve training stability. Only the accelerometer sensor data were used for this setup, excluding other modalities such as the gyroscope or the magnetic field.

\begin{table}[H]
\centering
\begin{tabular}{|c|c|}
\hline
\textbf{Subject ID} & \textbf{Windows} \\ \hline
0 & 1677 \\
1 & 1513 \\
2 & 1780 \\
3 & 1602 \\
4 & 1920 \\
5 & 1687 \\
6 & 1490 \\
7 & 1905 \\
8 & 1753 \\
9 & 1643 \\
10 & 1720 \\
11 & 1660 \\
12 & 1708 \\
13 & 1420 \\
14 & 1735 \\
\hline
\end{tabular} \hspace{5mm}
\begin{tabular}{|c|c|}
\hline
\textbf{Label} & \textbf{Windows} \\ \hline
walking & 3771 \\ 
running & 4078 \\
sitting & 3739 \\
standing & 3697 \\
lying & 3748 \\
climbing up & 3107 \\
climbing down & 2502 \\
jumping & 571 \\
\hline
\end{tabular}
\caption{RealWorld2016 number of windows per subject (left) and per activity (right).}\label{tab:RW2016}
\end{table}


Table \ref{tab:RW2016} show windows count per subject and activity, respectively. 

\subsection{Model and Framework}
We used a Convolutional Neural Network (CNN) as a model for both the traditional centralized learning and federated learning. The model consists of 192 convolutional filters followed by a max-pooling layer for feature reduction. The extracted features were then passed through a fully connected dense layer with 1,024 units, with a dropout rate of 0.5 applied to prevent overfitting. The training process used a learning rate of 0.01, a batch size of 32, and was run for 5 epochs per local training round. The overall training was performed in a federated learning framework over 200 rounds, where each client trained locally before model aggregation, ensuring privacy-preserving distributed learning.

To carry out our exploration, we used Flower \cite{beutel2022flowerfriendlyfederatedlearning}, a novel end-to-end federated learning framework that enables a more seamless transition from experimental research in simulation to system research on a large cohort of real edge devices. Flower offers individual strength in both areas (simulation and real-world devices), as well as the ability for experimental implementations to migrate between the two extremes as needed during exploration and development.

In particular, for the experiment considered in this work, which implements a fully synchronous federated learning setting, the Flower strategy parameters were configured so that all clients participate in the training process and the overall behavior is deterministic.

In addition, in order to obtain the desired results, metrics were tracked at the local, global, and personalized levels, and an explicit saving of both global and personalized model weights was performed at the end of the training rounds.

This setup made it possible to analyze the metrics described in Section \ref{sec:data-partitioning} and in Section \ref{sec:additional-characterization}, compare the results obtained on the training and validation sets, and assess the effects of the class removal strategy based on the round robin scheme.

\subsection{Design of Experiments}
To obtain the desired results, the analysis was structured to enable a comparison among different learning paradigms, namely \emph{centralized}, \emph{local}, and \emph{federated learning}, thereby assessing both model generalization and personalization capabilities. In particular, each experiment described below was conducted by running the training process with five different random seeds, and the performance metrics, accuracy and loss, are reported as the mean across seeds together with the corresponding 95\% confidence interval (CI95).

\subsubsection{\textbf{Centralized Model}}
A centralized training procedure is performed on the entire dataset, obtained by merging the 15 individual training sets corresponding to each client, thus allowing the use of data from all subjects. The training is carried out for 1000 epochs, derived from the total number of iterations used in the federated setting (200 rounds with 5 local epochs each).
The validation set, obtained by merging the 15 individual testing sets corresponding to each client, is used to monitor the training process and to activate early stopping mechanisms, configured with a patience value of 10, based on the validation accuracy achieved during training, with the aim of preventing overfitting.

\subsubsection{\textbf{Local Models}}
In this scenario, there is no single centralized model; instead, 15 separate training processes are performed, one for each client, with each model trained exclusively on its own local training dataset.
In particular, local training is performed using the same hyperparameters adopted for centralized training, also applying an early stopping strategy in this case. Subsequently, it is possible to evaluate the performance of the individual models using as evaluation sets both the corresponding local test set of each client (\emph{L-L}) and the global test set (\emph{L-G}).

\subsubsection{\textbf{Federated Learning}}
A federated learning system based on the FedAvg algorithm is implemented. Specifically, each client locally trains a CNN-based model using its own local data. 
The federated training follows a round-based scheme, where in each round every client performs five local training epochs. At the end of each round, the locally updated model weights are aggregated and assigned to the global model, which is reconstructed at every round and subsequently used both for the next training iteration and for global evaluation.

Regarding validation, several metrics are logged into a CSV file at each round and for each client. These include personalized accuracy (\emph{PF-L}), federated (i.e. aggregated) accuracy evaluated on local test sets (\emph{F-L}) and on global test sets (\emph{F-G}), along with the corresponding loss values. In addition, we also evaluate the generalization accuracy of personalized models, by testing all of them on the global dataset (\emph{PF-G}).

In this case, with respect to the federated model (\emph{F-}), it refers to the globally aggregated federated model, which is obtained by aggregating the client-updated weights round by round. This follows directly from the definition of federated learning, since one of its main objectives is to preserve client privacy by avoiding any form of data sharing (which would occur when using a global training set), and instead exchanging only model parameters.

Nevertheless, a global training set is still constructed in the same manner as for the centralized model, but it is used exclusively for the statistical evaluation of accuracy. As for the test set, it is accordingly formed by aggregating the remaining 20\% portions of the clients’ local test sets into a unified dataset.


\subsubsection{\textbf{Round Robin Class Exclusion}}
Regarding experiments involving class removal through the round robin strategy, the same experimental setup described in the previous sections is adopted, preserving the same validation protocols. This strategy is introduced to assess the robustness and learning capability of the models in the presence of missing classes at the client level, as well as to evaluate the effective generalization capability of the models within a federated learning framework.

As a consequence, given a total of 15 clients and 8 activity classes from the RealWorld2016 dataset, each class is removed from approximately the same number of clients. This guarantees a balanced class absence across participating clients, preventing biases or distortions toward specific activities and enabling a fair comparison of model behavior under missing-class conditions.

In particular, in addition to the experiments described above, a specific client-wise evaluation was performed on the corresponding removed classes. After excluding the assigned class from each client during the model pre-training phase, the samples belonging to the removed classes were stored in a separate data structure, so that they could be used both as a Local Dropped test set \textit{(-LD)} for each client and to perform, at the end of the respective training procedures, a validation of the final centralized model \textit{(C-LD\_RR)}, the local models, whose performance resulted equal to zero for all clients, as the evaluated class had never been observed during training \textit{(L-LD\_RR)}, the personalized models \textit{(PF-LD\_RR)}, and the final federated model \it{(F\-LD\_RR)}.

Thus, the validation on the removed classes highlights the trade-off between global generalization and client-specific personalization, enabling the evaluation of the effectiveness of personalization within the federated setting, even under heterogeneous data regimes.

\section{Experimental Results and Discussion}\label{sec:results}
This section reports the performance of centralized, local, federated, and personalized models in terms of classification accuracy on both local and global datasets.
According to the methodology and the experimental setup described in Section \ref{sec:method} and \ref{sec:expsetup}, respectively, we summarize the list of reported results as follows:
\begin{itemize}
    \item Validation of Centralized Model on Global Dataset (C-G and C-G\_RR) and on Local Datasets (C-L and C-L\_RR)
    \item Validation of Local Models on Global Dataset (L-G and L-G\_RR) and on Local Datasets (L-L and L-L\_RR)
    \item Validation of Federated Model on Global Dataset (F-G and F-G\_RR) and on Local Datasets (F-L and F-L\_RR)
    \item Validation of Personalized Models on Global Dataset (PF-G and PF-G\_RR) and Local Datasets (PF-L and PF-L\_RR)
    \item Validation of Centralized Model on Local Dropped Class (C-LD\_RR)
    \item Validation of Federated Model on Local Dropped Class (F-LD\_RR)
    \item Validation of Personalized Models on Local Dropped Class (PF-LD\_RR)
\end{itemize}

\begin{figure}[H]
    \centering
    \includegraphics[width=0.9\linewidth]{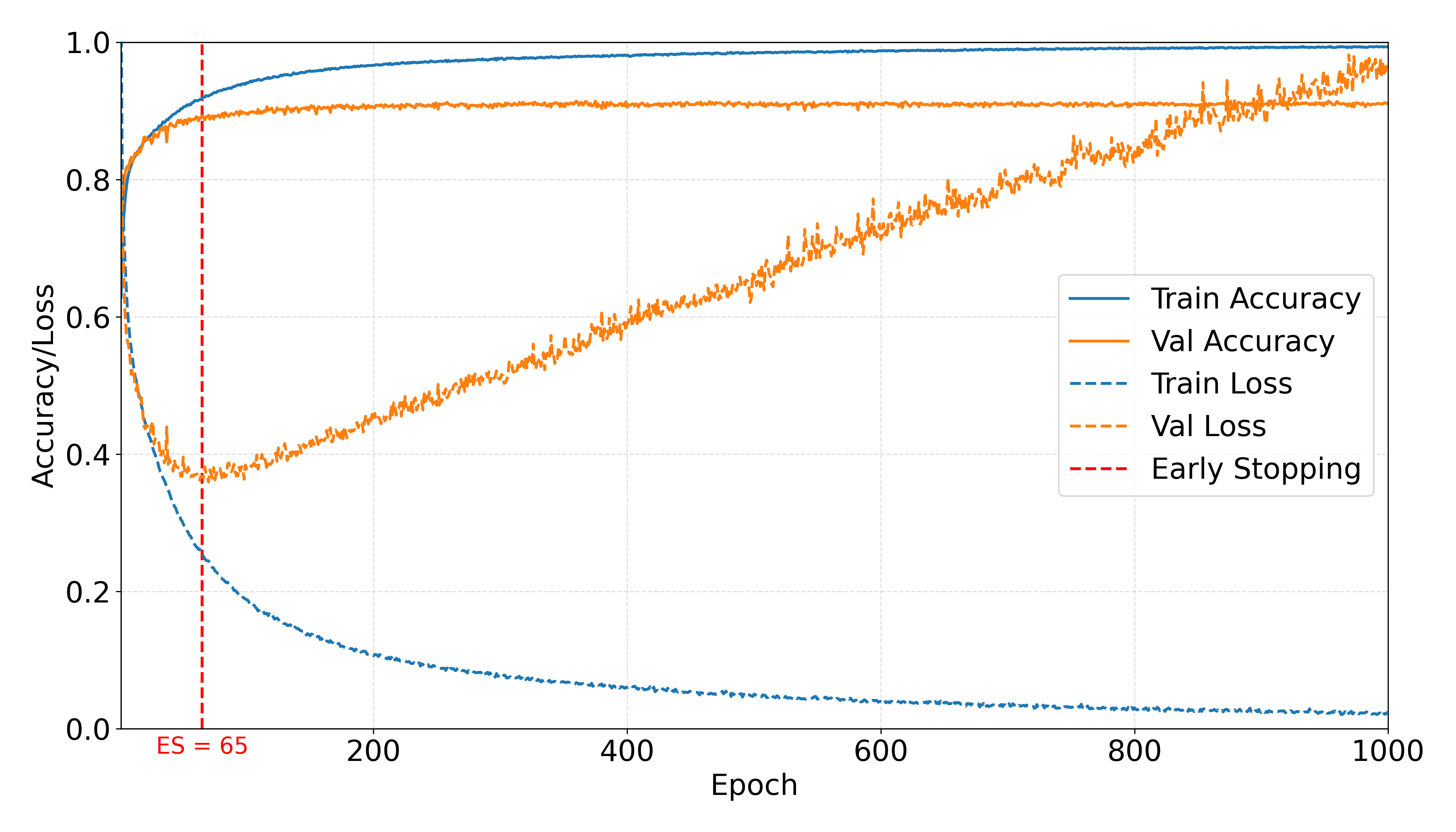}
    \caption{Training and validation accuracy and loss without Early Stopping in C-G.}
    \label{fig:CGwoutES}
\end{figure}

\begin{figure}[H]
    \centering
    \includegraphics[width=0.9\linewidth]{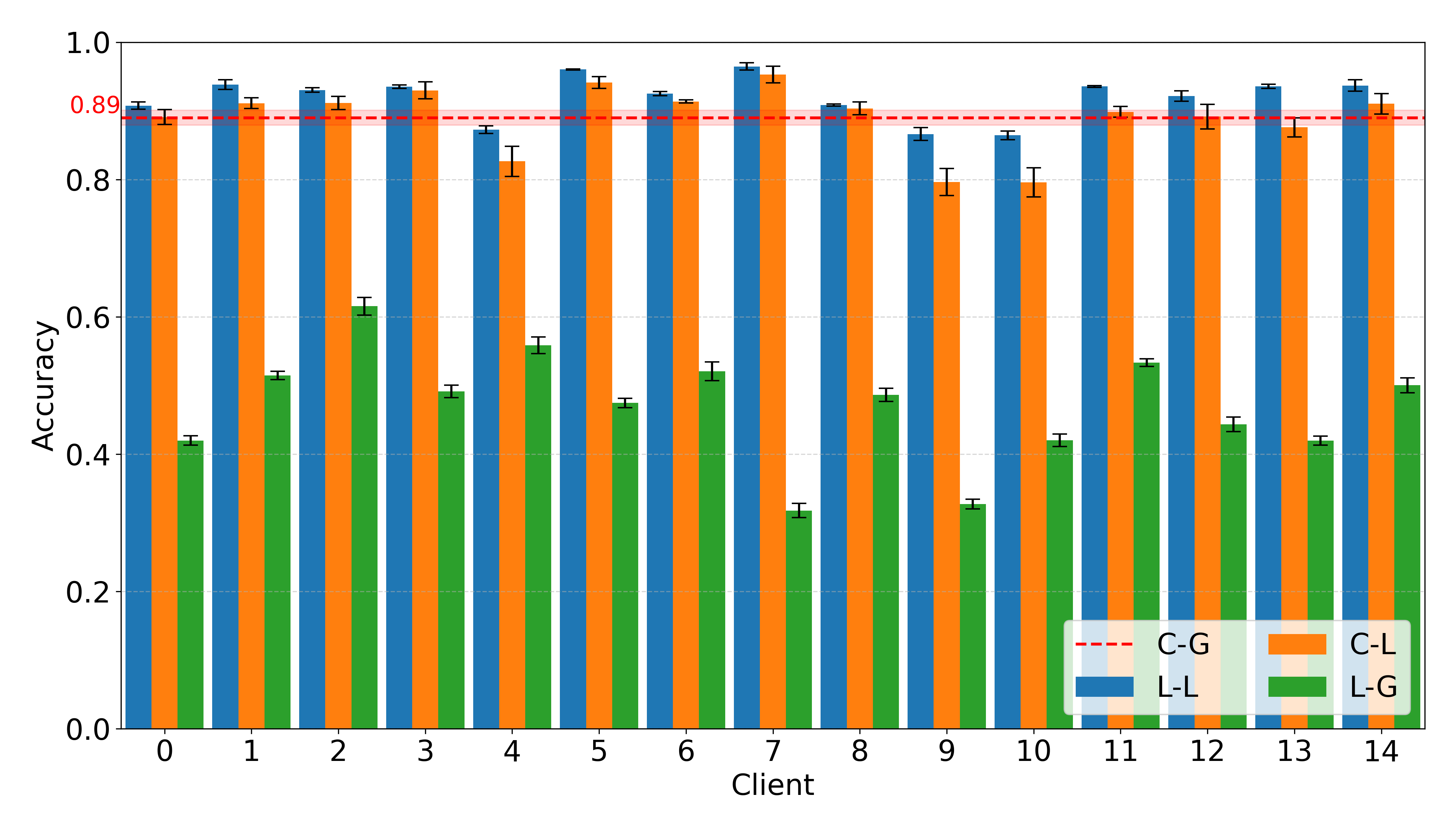}
    \caption{Accuracy of local (L-) and centralized (C-) models, with early stopping at epoch 65, evaluated on global (-G) and local (-L) validation sets.}
    \label{fig:centralizedVSlocal}
\end{figure}

\subsection{Centralized and Local Scenario}
We start by analyzing the centralized and local scenarios to assess the performance in the absence of a federation, i.e., without a collaborative approach.

Figure \ref{fig:CGwoutES} shows the training and validation accuracy, and loss without any early stopping mechanism evaluated for the global scope (C-G). The goal of this evaluation is twofold: on the one hand, it assesses the performance of the traditional centralized learning, scoring an accuracy of 0.89. On the other hand, it allows us to identify the number of epochs at which it is desirable to stop training, thus avoiding the risk of overfitting. According to the early stopping patience parameter set to 10, to avoid overfitting, the training stopped at epoch 65. All the results reported hereafter were obtained with early stopping.

We then characterize the performance of each client's local model, validated in both the local (L-L) and global (L-G) scope. Those performances are compared with those of the centralized model in both scopes (namely, C-L and C-G). The results are reported in Figure \ref{fig:centralizedVSlocal}. We observe that, for each client, the local models (blue bars, L-L) achieve better accuracy than the centralized one (orange bars, C-L), because of the higher specialization of the models trained only on local data. However, the local models show their generalization limits, with accuracy between 0.3 and 0.6, when tested on global datasets (green bars, L-G), in comparison with the accuracy achieved by the centralized model (red dashed line, C-G), 0.89.


\subsection{Federated Scenario}
\begin{figure}[htb]
    \centering
    \includegraphics[width=0.9\linewidth]{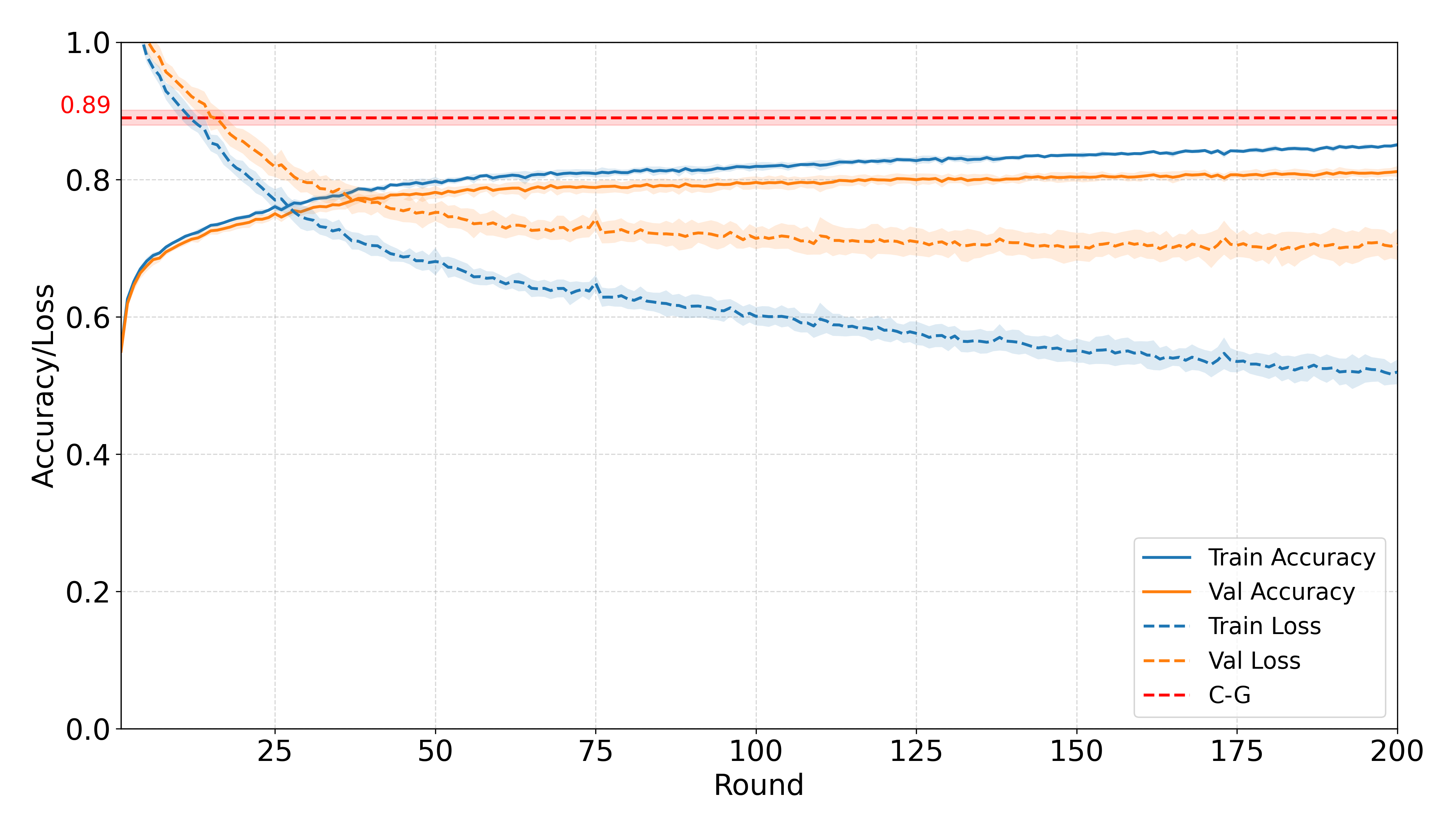}
    \caption{Training and validation accuracy and loss of the federated model evaluated on the global validation set (F-G) across 200 rounds.}
    \label{fig:fedavgGblVScentrGlb}
\end{figure}

Figure \ref{fig:fedavgGblVScentrGlb} shows the training and validation accuracy, and loss of the federated (FedAvg) model evaluated in the global scope (F-G) across 200 rounds. This is relevant to assessing the global accuracy that we compare with traditional centralized learning (red dashed line, C-G). As expected, the global accuracy F-G is slightly lower than traditional centralized accuracy C-G.

\begin{figure}[htb]
    \centering
    \includegraphics[width=0.9\linewidth]{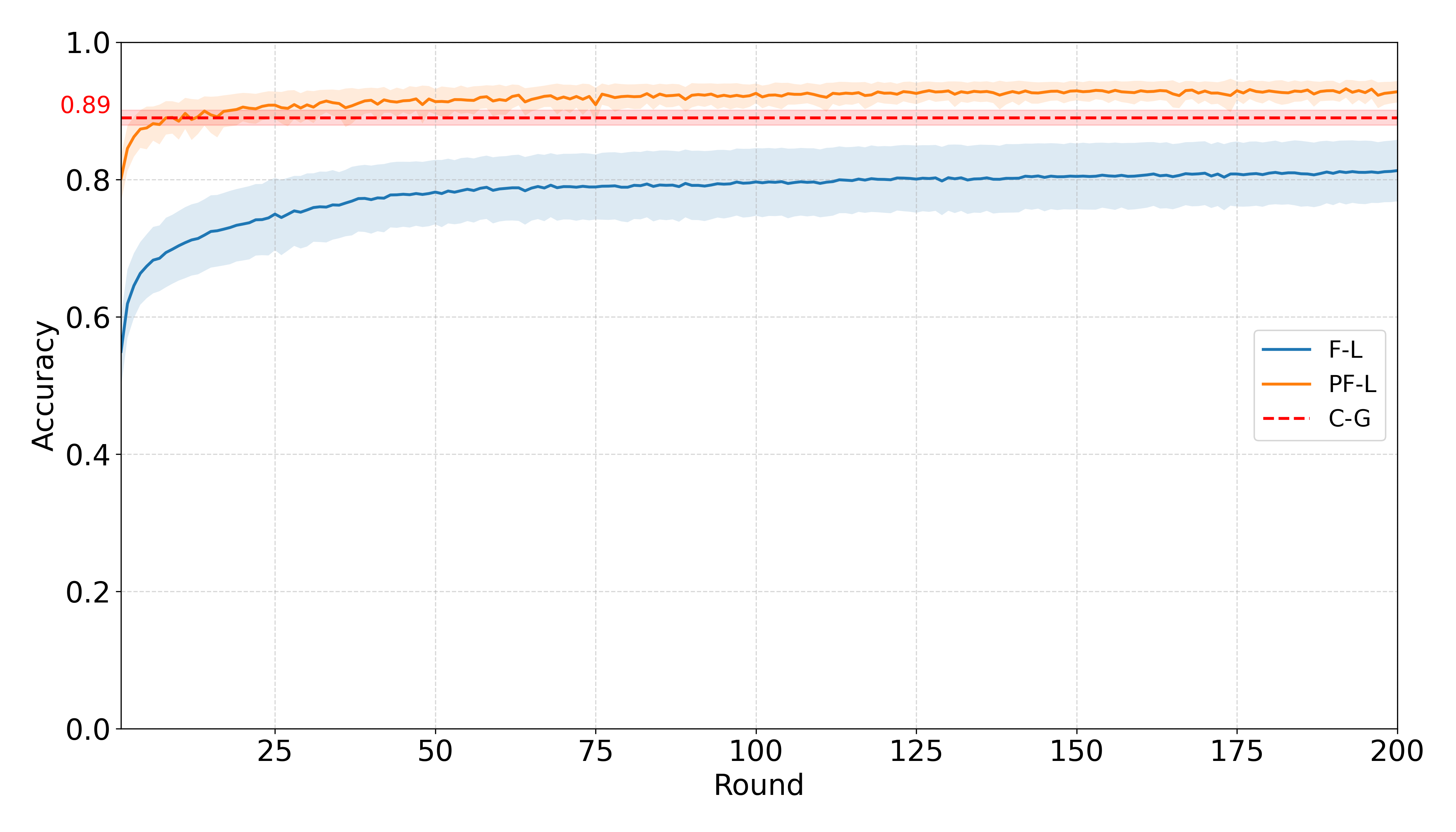}
    \caption{Mean of the Local validation accuracy of Federated (F-L) and Personalized Federated (PF-L) models across 200 rounds.}
    \label{fig:persfedavgPF-L}
\end{figure}

Figure \ref{fig:persfedavgPF-L} shows the mean of local validation accuracy of the federated model (blue line, F-L) and of the personalized model (orange line, PF-L). F-L performance on each local dataset is lower compared to PF-L, even if it increases round by round, as expected. However, after the local fine-tuning carried out by the clients, Figure \ref{fig:persfedavgPF-L} also shows the improved personalized accuracy, which increases round by round, as well. This is expected because each client needs to adapt to the evolving model adjustments.

\begin{figure}[htb]
    \centering
    \includegraphics[width=0.9\linewidth]{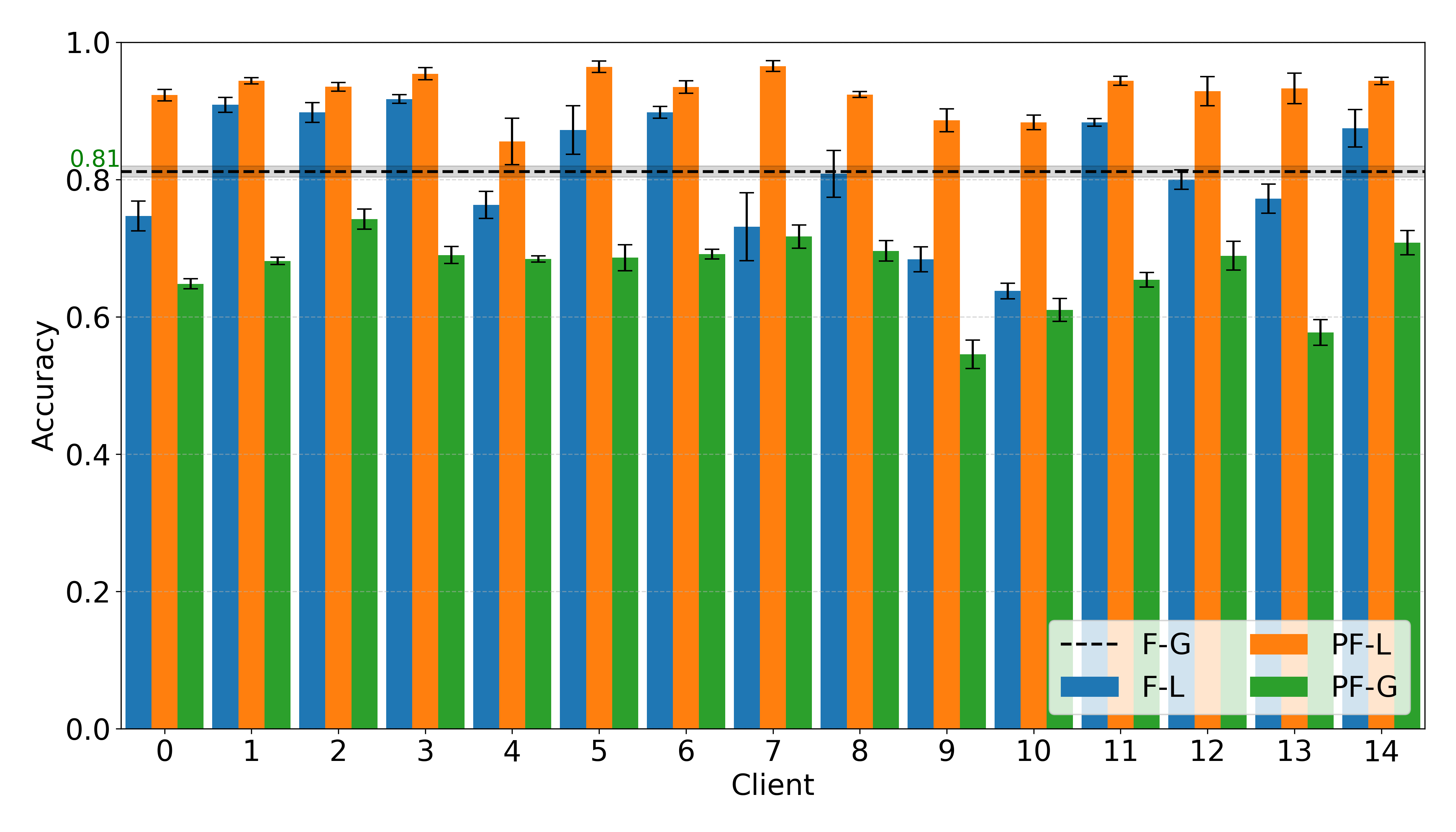}
    \caption{Accuracy of Federated (F-) and Personalized federated (PF-) models evaluated on local (-L) and global (-G) datasets.}
    \label{fig:federatedVSpersonalized}
\end{figure}

Figure \ref{fig:federatedVSpersonalized} shows the performance of the federated model and that of the personalized models, both validated on the local (F-L) and global test sets (F-G). The server scores a generalization capability of 0.81 on the global test set (F-G), which is 0.08 lower than the traditional centralized learning. The clients score a generalization capability that ranges between 0.55 and 0.7 (green bars, PF-G), and a personalization capability higher than 0.8 (orange bars, PF-L). Comparing the results obtained in Figure \ref{fig:centralizedVSlocal} (the non-federated scenario) with those of Figure \ref{fig:federatedVSpersonalized} (the federated scenario), the central server model loses in terms of generalization capability ($F-G < C-G$), but clients take advantage of the federation, improving their ability to generalize over unseen data ($PF-L > L-G$).

\subsection{Round Robin Activity Class Exclusion}
When implementing the policy of activity class exclusion, interesting performance characteristics were observed with respect to personalization and generalization capabilities.

\subsubsection{Centralized and Local Scenario}
\begin{figure}[htb]
    \centering
    \includegraphics[width=0.9\linewidth]{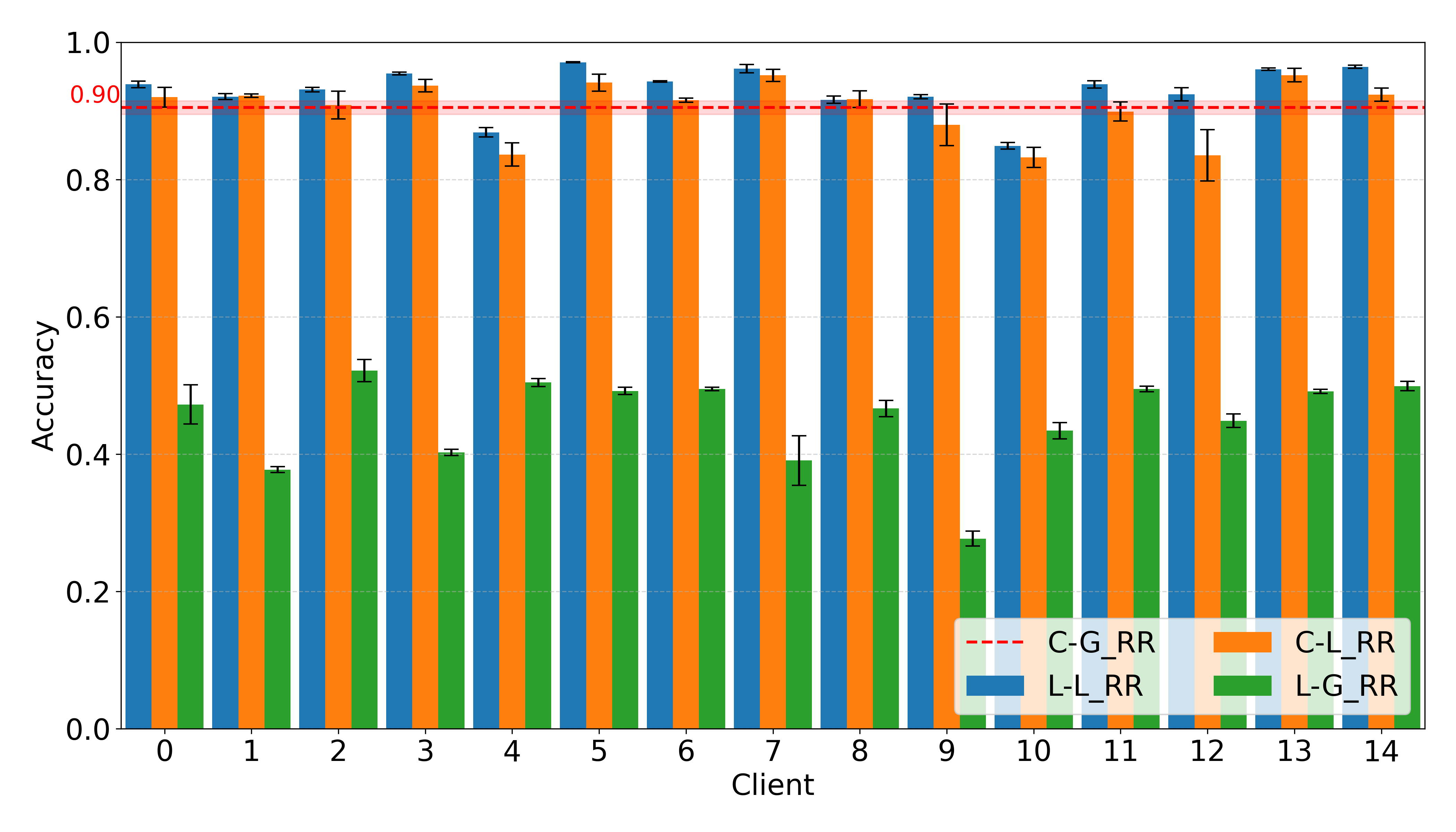}
    \caption{Accuracy of local (L-) and centralized (C-) models, with early stopping at epoch 65, evaluated on global (-G) and local (-L) validation sets with round robin activity removal (\_RR).}
    \label{fig:centrVSlocalRR}
\end{figure}

Considering the early stopping mechanism, Figure \ref{fig:centrVSlocalRR} shows that, for each client, the local models (blue bars, L-L\_RR) achieve better accuracy than the centralized one (orange bars, C-L\_RR), because of the higher specialization of the models trained only on local data. However, the local models experience generalization limits, with accuracy between 0.25 and 0.5, when tested on global datasets (green bars, L-G\_RR), in comparison with the accuracy achieved by the centralized model (red dashed line, C-G\_RR), 0.90 .

\subsubsection{Federated Scenario}
\begin{figure}[htb]
    \centering
    \includegraphics[width=0.9\linewidth]{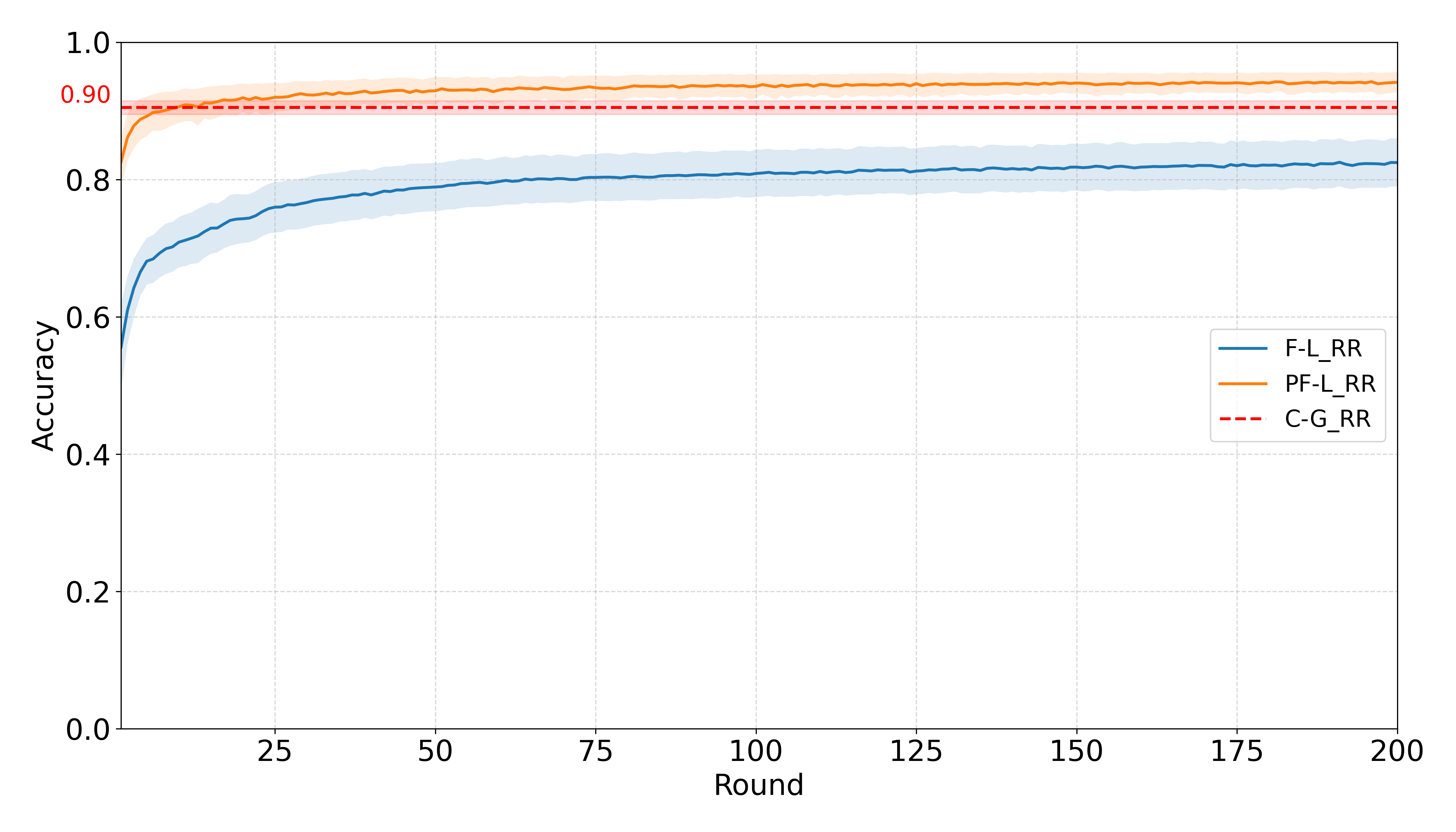}
    \caption{Mean of the Local validation accuracy of Federated (F-L\_RR) and Personalized Federated (PF-L\_RR) models across 200 rounds in case of round robin activity removal.}
    \label{fig:persfedavgPF-L-RR}
\end{figure}

The impact of federation is similar to the results obtained without activity class exclusion with respect to clients’ personalization capabilities. Indeed, Figure \ref{fig:persfedavgPF-L-RR} shows the mean of local validation accuracy of the federated model (blue line, F-L\_RR) and of the personalized model (orange line, PF-L\_RR). F-L\_RR performance on each local dataset is lower compared to PF-L\_RR , even if it increases round by round, as expected. However, after the local fine-tuning carried out by the clients, Figure \ref{fig:persfedavgPF-L-RR} also shows the improved personalized accuracy, which increases round by round, as well. This is expected because each client needs to adapt to the evolving model adjustments.

\begin{figure}[htb]
    \centering
    \includegraphics[width=0.9\linewidth]{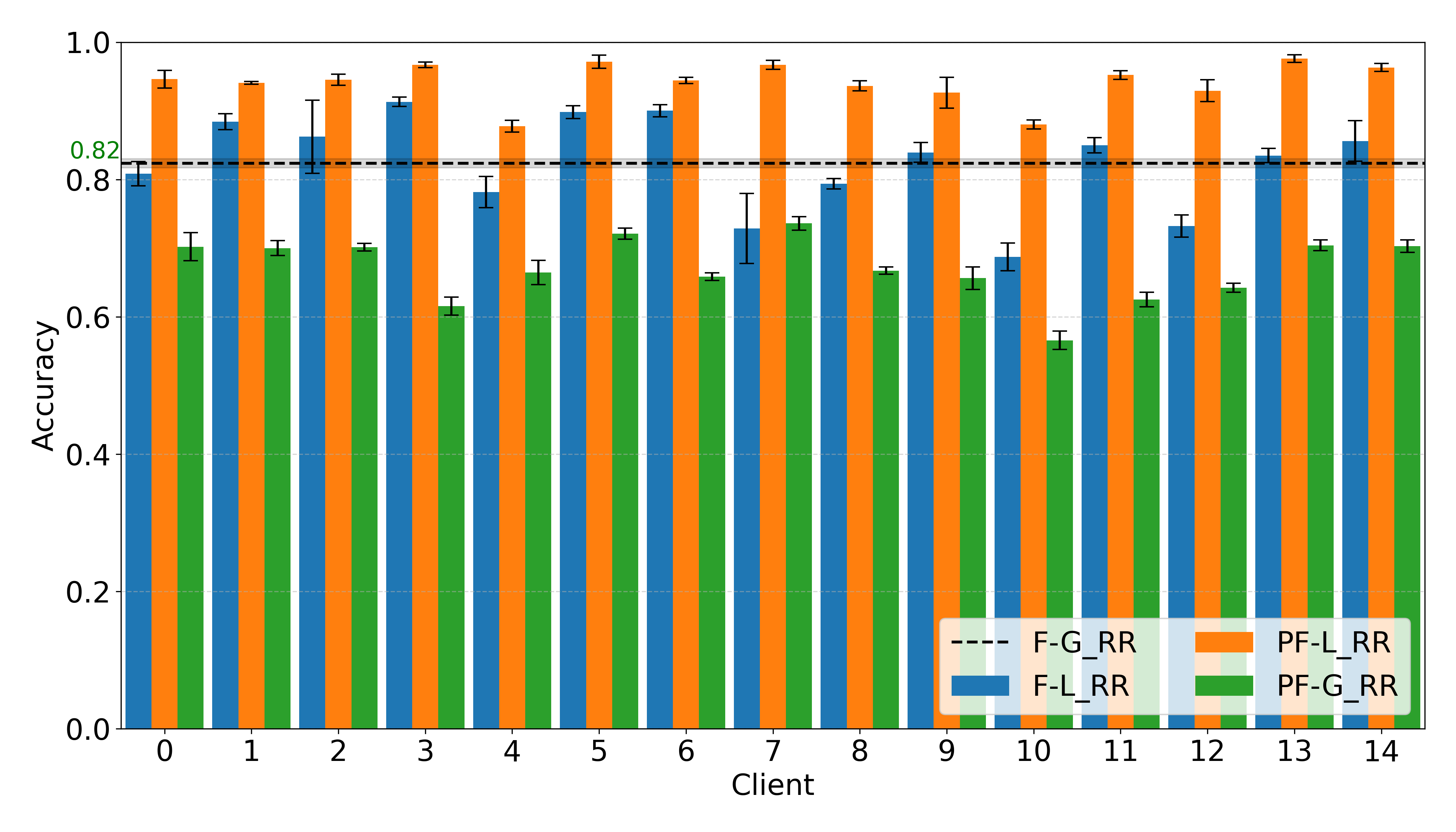}
    \caption{Accuracy of Federated (F-) and Personalized federated (PF-) models evaluated on local (-L) and global (-G) datasets in case of round robin activity removal (\_RR).}
    \label{fig:federatedVSpersonalizedRR}
\end{figure}

Figure \ref{fig:federatedVSpersonalizedRR} shows the performance of the federated model and that of the personalized models, both validated on the local (F-L\_RR) and global test sets (F-G\_RR). The server scores a generalization capability of 0.82 on the global test set (F-G\_RR), which is 0.09 lower than the traditional centralized learning. The clients score a generalization capability that ranges between 0.55 and 0.7 (green bars, PF-G\_RR), and a personalization capability higher than 0.8 (orange bars, PF-L). Comparing the results obtained in Figure \ref{fig:centrVSlocalRR} (the non-federated scenario) with those of Figure \ref{fig:federatedVSpersonalized} (the federated scenario), the central server model loses in terms of generalization capability ($F-G\_RR < C-G\_RR$), but clients take advantage of the federation, improving their ability to generalize over unseen data ($PF-L\_RR > L-G\_RR$).

\begin{figure}[htb]
    \centering
    \includegraphics[width=0.9\linewidth]{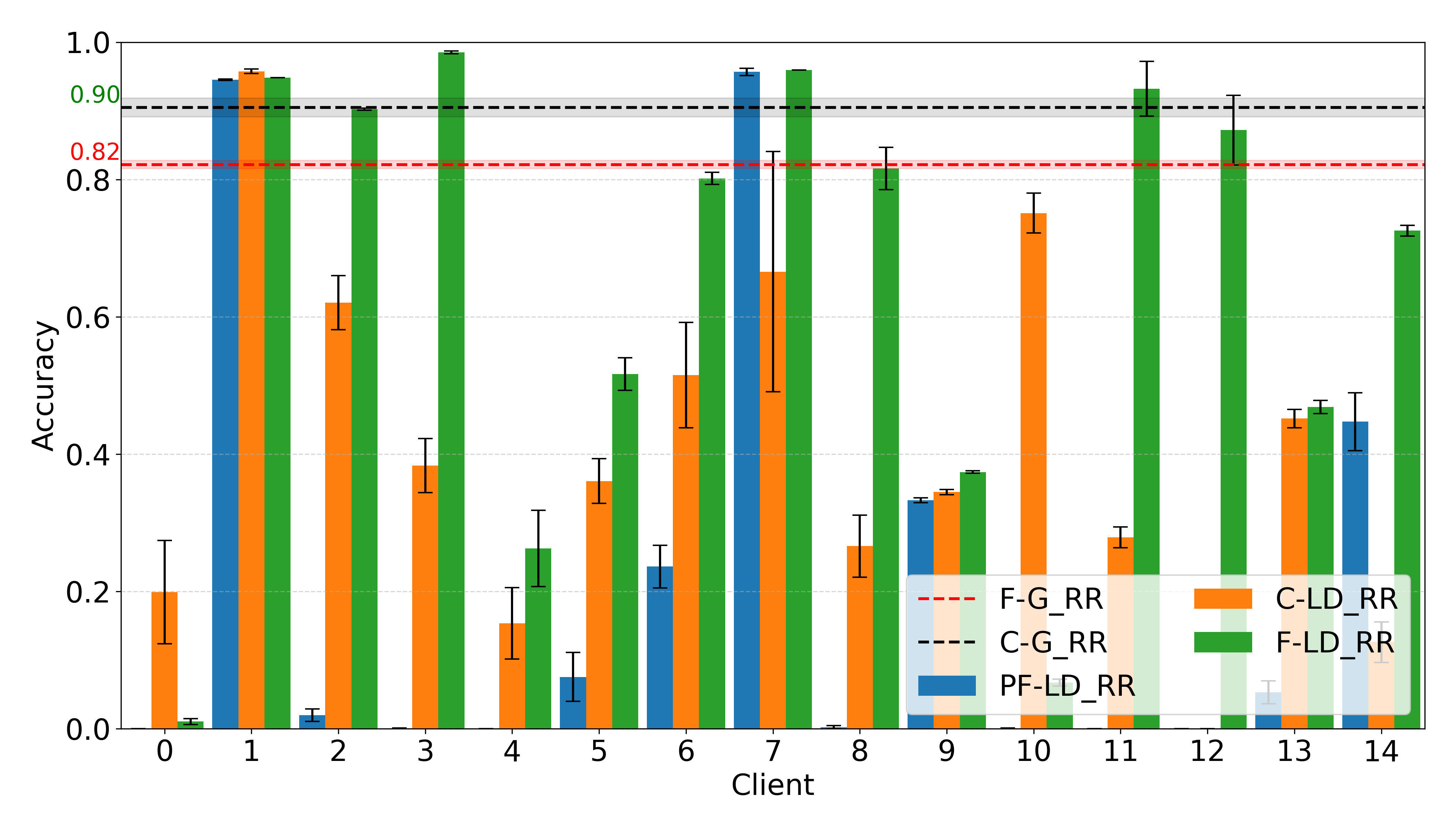}
    \caption{Generalization accuracy of all the models (C\_RR, F\_RR, and PF\_RR) evaluated globally (-G) and locally to each client, using as local validation sets the activities removed from the training set (-LD).}
    \label{fig:allmodelsRR}
\end{figure}

Figure \ref{fig:allmodelsRR} shows the results of a stress test evaluating local generalization performance in a scenario where all the models are used to evaluate unseen subject-class pairs. In fact, the test set used in this case is composed only of those data that were removed from the training and test sets by the round robin class exclusion procedure. Hence, the personalized models of each client were asked to recognize classes that they had never seen before, while the centralized and aggregated models were asked to recognize data belonging to unseen subject-class pairs. Needless to say, local models were not used in this experiment because they have no means to recognize unseen classes. Please notice that the horizontal lines are reported only for comparison, since they refer to the performance achieved by aggregated and centralized models on the global RR test set, as reported and discussed in the previous two graphs.

In federated learning, the aggregated model (F-LD) outperforms the personalized one (PF-LD) for all the clients. This is expected, but not obvious, suggesting that the aggregated models before local personalization can be used by the clients to screen unusual classes. It is also worth noticing that, in many cases, the performance obtained on this extreme case by the federated model overcomes its average performance on the test set.

The centralized model (C-LD) performs much worse, confirming the superior generalization capability of the federated model, even if in many cases it overcomes the performance of personalized models.


\section{Conclusions} \label{sec:conclusion}
In this study, we designed and tested different scenarios to study how the federated learning algorithm FedAvg behaves on the HAR task compared to the local and centralized paradigms to gain insight into the personalization and generalization capabilities. To further investigate these abilities, we not only propose further evolution metrics, but we also designed and tested variations in class distribution across clients. Although the federated FedAvg approach confirms a higher degree of personalization capabilities while keeping a high degree of generalization with respect to the traditional centralized learning, this result is not so obvious under stressful conditions, such as varying class distribution over clients. Indeed, such a scenario requires further investigation to gain much more insights about personalization and generalization capabilities. We plan to test our activity class exclusion protocol on more extensive datasets from different users.

\section*{Acknowledgment}
This work was partially funded by MIMIT, under FSC project "Pesaro CTE SQUARE", CUP D74J22000930008.

\bibliographystyle{ieeetr}
\bibliography{biblio.bib}

\end{document}